\documentclass[11pt,a4paper]{article}
\usepackage[hyperref]{naaclhlt2018}
\usepackage{times}
\usepackage{latexsym}
\usepackage{booktabs}
\usepackage{linguex}
\usepackage{pbox}
\usepackage{amsmath}
\usepackage[shortcuts]{extdash}

\usepackage{url}
\usepackage[]{todonotes}   
\usepackage{multirow}
\hypersetup{draft}

\usepackage{array}
\newcolumntype{$}{>{\global\let\currentrowstyle\relax}}
\newcolumntype{^}{>{\currentrowstyle}}

\aclfinalcopy 
\def\aclpaperid{91} 

\title{Verb Argument Structure Alternations in\\Word and Sentence Embeddings}

\author{  
  Katharina Kann\thanks{\; The first three authors contributed equally and are listed in alphabetical order.} , Alex Warstadt$^*$, Adina Williams$^*$ and Samuel R. Bowman\\ 
  New York University, USA\\
  \texttt{\{kann, warstadt, adinawilliams, bowman\}@nyu.edu} 
}

\date{}

\begin{document}
\maketitle
\begin{abstract}
Verbs occur in different syntactic environments, or frames. We investigate whether artificial neural networks encode grammatical distinctions necessary for inferring the idiosyncratic frame-selectional properties of verbs.
We introduce five datasets, collectively called FAVA, containing in aggregate nearly 10k sentences labeled for grammatical acceptability, illustrating different verbal argument structure alternations. 
We then test whether models can distinguish acceptable English verb--frame combinations from unacceptable ones using a sentence embedding alone.
For converging evidence, we further construct LaVA, a corresponding word-level dataset, and investigate whether the same syntactic features can be extracted from word embeddings.
Our models perform reliable classifications for some verbal alternations but not others, suggesting that while these representations do encode fine-grained lexical information, it is incomplete or can be hard to extract.
Further, differences between the word- and sentence-level models show that some information present in word embeddings is not passed on to the downstream sentence embeddings.
\end{abstract}

\setlength{\Exlabelsep}{0em}
\setlength{\SubExleftmargin}{1.5em}
\section{Introduction}

Artificial neural networks (ANNs) are powerful computational models that are able to implicitly learn syntactic and semantic features necessary for a variety of natural language tasks. These empirical results raise a deeper scientific question: to what extent do the features learned by ANNs resemble the linguistic competence of humans? 

\setlength{\tabcolsep}{5pt}
\begin{table*}[ht] 
    \footnotesize 
  \begin{tabular}{clll}
    \toprule
    Verb Frame & \multicolumn{3}{l}{ Example Sentences} \\
    \toprule
    \multirow{2}{*}{}{Caus.} & {Jessica \textbf{dropped} the vase.} & {Jessica blew the bubble.}  \\
     {Inch.} & {The vase \textbf{dropped}.} & {*The bubble blew.} \\
    \midrule
    \multirow{2}{*}{}{Dative-Prep.} & {Liz \textbf{gave} a gift to the boy.} & {Liz administered a test to the kid.} & {*Liz charged \$50 to Jon.} \\
   {Dative-2-Obj.} & {Liz \textbf{gave} the boy a gift.} & {*Liz administered the kid a test.} & {Liz charged Jon \$50.} \\
    \midrule
    \multirow{2}{*}{}{Spr.-Lo.-\textit{with}} & {Sue \textbf{loaded} the truck with wood.}& {Sue coated the deck with paint.} & {*Sue swept the bin with sand.}\\
   {Spr.-Lo.-Loc.}  & {Sue \textbf{loaded} wood onto the truck.}& {*Sue coated paint on the deck.} & {Sue swept sand into the bin.}\\
    \midrule
    \multirow{2}{*}{}{{no-\textit{there}}} & {Fear \textbf{remained} in my mind.} & {A girl focused on the quiz.} \\
    {\textit{there}} & {There \textbf{remained} fear in my mind.} & {*There focused on the quiz a girl.} \\
    \midrule
    \multirow{2}{*}{}{U.-Obj.-Refl.} & {Ada \textbf{clapped} her hands.} & {Ada permed her hair.} & {*Ada exercised herself.} \\
    {U.-Obj.-No-Refl.} & {Ada \textbf{clapped.}} & {*Ada permed.} & {Ada exercised.} \\
    \bottomrule
  \end{tabular}
  \caption{\label{tab:FrameExamples} Examples from each verb frame in the dataset. Bolded verbs evoke both verb frames; other verbs evoke only one. Transitive verb frames include: Causative, \textsc{Spray--Load} \textit{with}, \textsc{Spray--Load} locative, \textsc{Understood-Object} reflexive. Intransitive verb frames include: Inchoative, no-\textit{there} (with locative adjunct), \textit{there} (with locative adjunct), and \textsc{Understood-Object} no-reflexive. 2-obj. class includes a ditransitive frame and a prepositional dative frame.}
\end{table*}

Studying the linguistic competence of ANNs, in addition to its intrinsic value for model evaluation, can help resolve outstanding scientific questions in linguistics about the role of prior grammatical bias in human language acquisition. \newcite{chomsky-65} suggests that the acquisition of rich grammatical distinctions is facilitated by an innate universal grammar (UG), which imparts specific grammatical knowledge to the learner. This proposal crucially depends on the \textit{poverty of the stimulus} argument, which holds that the acquisition of certain linguistic features by purely domain-general data-driven learning should not be possible \cite{clark-11}. 
Studying the ability of low-bias learners like ANNs to acquire specific grammatical knowledge can provide evidence relevant to this argument.

In this work, we evaluate ANNs' treatment of verbs; verbs contribute to the overall meaning of sentences by encoding information about how entities are related to, and participate in, events. Concretely, we investigate if ANNs acquire the specific grammatical distinctions necessary for inferring the frame-selectional properties of verbs. Cross-linguistically, the lexical entry of a verb is associated with a set of syntactic contexts or \textit{syntactic frames} in which it can appear. This information is lexically idiosyncratic, i.e., even verbs that are intuitively very similar in meaning may vary as to which syntactic frames they can appear in:
\ex.\label{ex:1}
\a.\label{ex:1:a}Sharon \textbf{sprayed} water on the plants. 
\b.\label{ex:1:b}Sharon \textbf{sprayed} the plants with water. 
\b.\label{ex:1:c}Carla \textbf{poured} lemonade into the pitcher. 
\b.\label{ex:1:d}*Carla \textbf{poured} the pitcher with lemonade.\footnote{In this paper, stars mark ungrammatical sentences.}

Certain verbs, e.g., \textit{spray}, select multiple related frames and are therefore known as \textit{alternating verbs}. In contrast, other semantically similar verbs, e.g., \textit{pour}, select only a single frame and are thus not alternating. Information about whether a given verb alternates (as well as which frames it can appear in) has been described and classified in several verb lexica \citep{grishman1994, baker1998, fillmore2003, kipper2005, kipper2006}. Knowledge about verb frames and their alternations is part of a human speaker's linguistic competence, and as such, should potentially be learned by ANNs.

We present two datasets and two experiments that compare ANNs' knowledge of verb frame alternations at the word level and the sentence level, respectively. First, we ask if a verb's word embedding can be used to predict which frames that verb can licitly appear in. We construct a dataset of verbs, the \textbf{L}exic\textbf{a}l \textbf{V}erb--frame \textbf{A}lternations dataset (LaVA), based on \newcite{levin1993}, and train a multi-class classifier to identify the licit syntactic frames associated with a verb from its word embedding alone (if successful, the classifier should be able to determine, e.g., that \textit{sprayed} alternates and can appear in sentences with \textit{with}-alternants like \ref{ex:1:b}, but that \textit{poured} cannot \ref{ex:1:d}). 

Second, we ask whether sentence embeddings encode the frame-selectional properties of their main verb. The main verb's frame-selectional properties have consequences for
grammaticality at the sentence level; to give an example, \ref{ex:1:d} is not grammatical, because \textit{poured} cannot participate in this frame alternation. To exploit this, we semi-automatically generate sentences 
in such a way to ensure that the main verb's frame alternation information is the only information determining the (un)grammaticality of the sentence. For a portion of the sentences, the main verb can participate in a given verb frame alternation, and for another portion it cannot; if the main verb cannot participate in the alternation, then one of the sentences in the pair will be ungrammatical. Using this dataset, the \textbf{F}rames and \textbf{A}lternations of \textbf{V}erbs \textbf{A}cceptability dataset (FAVA), we train a binary classifier to judge the acceptability of sentences containing verbs in various syntactic contexts using the sentence embeddings alone. 

We find that verb frame information is extractable from both word embeddings and sentence embeddings, but that these two complementary methods differ in performance. The LaVA and FAVA datasets are available under \href{https://nyu-mll.github.io/CoLA}{\url{https://nyu-mll.github.io/CoLA}} for future research and model evaluation. 

\begin{table*}[ht] 
  \setlength{\tabcolsep}{6.5pt}
  \small
  \centering
  \begin{tabular}{ l  cc  cc  cc  cc  cc }
    \toprule
    Levin class & \multicolumn{2}{c }{\textsc{Caus.--Inch.}} & \multicolumn{2}{c }{\textsc{Dative}} & \multicolumn{2}{c }{\textsc{Spray--Load}} & \multicolumn{2}{c }{\textsc{\textit{there}-Insertion}} & \multicolumn{2}{c }{\textsc{Understood-Object}} \\
     & Inch. & Caus. & Prep. & 2-Obj. & \textit{with} & Loc. & no-\textit{there} & \textit{there} & Refl. & No-Refl.\\
    \midrule
    Positive & 70 & 120 & 63 & 72 & 90 & 81 & 50 & 145 & 11 & 81  \\
    Negative & 140 & (0) & 356 & 405 & 220 & 229 & 185 & (0) & 466 & 396 \\
    \midrule
    Total & 210 & 120 & 419 & 477 & 310 & 310 & 235 & 145 & 477 & 477 \\
    \bottomrule
  \end{tabular}
  \caption{\label{tab:overview}Overview of the lexical dataset. ``Positive'' refers to the number of verbs that evoke each frame (i.e., will yield a grammatical sentence) and ``negative'' refers to the number of verbs which do not evoke those frames (i.e., will yield an ungrammatical sentence). Causative and \textit{there} sentence frames have no negative examples (i.e., every verb participating in the alternation can instantiate these frames).
  } 
\end{table*}

\section{Verb Frame Alternations}\label{sec:FrameAlternations}

The lexical meaning of each verb includes a description of an event and how entities participate in it \cite{fillmore1966,fillmore2003}, and this information is present for the various syntactic frames associated with each verb. To determine whether our ANNs encode this information, we select five verb frame alternations from \citet{levin1993}; the verb frames which comprise each alternation vary either in the number of arguments they can take,  
in the order in which the arguments appear, 
or in both. 
Examples are given in Table \ref{tab:FrameExamples}, and statistics are provided in Tables \ref{tab:overview} and \ref{tab:corpuscount}. 

To give an example, in \ref{ex:1:a}, there is an event of \textit{spraying} in which \textit{Sharon} is the main actor (often referred to as \textit{agent}), \textit{the plants} is the  entity affected by the event (i.e., the \textit{patient}), and \textit{water} is the entity used in the event (i.e., the \textit{instrument} or \textit{theme}). In \ref{ex:1:a}, the verb frame of \textit{spray} has three roles, and they come in a specific order: the \textit{agent} is the subject, the \textit{instrument} is the object, and the \textit{patient} or \textit{location} is part of a prepositional phrase adjoined to the verb. Participants (e.g., \textit{Sharon} and \textit{water}) that are provided by the verb are called \textit{arguments} of the verb; the other argument \textit{the plants} is within a prepositional phrase and is therefore not provided by the verb.

Whether a verb can introduce a given number of arguments can affect its sentence-level grammaticality and is therefore of interest here. Verbs can be \textit{intransitive}, taking only one argument (e.g., \textit{dropped} in \textit{the vase dropped.}), 
\textit{transitive}, taking two arguments (e.g., \textit{dropped} in \textit{Jessica dropped the vase}), or \textit{ditransitive}, taking three arguments (e.g., \textit{gave}, in \textit{Liz gave the boy a gift}). 

Two different verb frames may be related by the addition or deletion of an argument (e.g., \textsc{causative-inchoative}), or by realizing the same arguments in a different syntactic configuration (e.g., \textsc{spray-load}; \ref{ex:1:a} and \ref{ex:1:b}). When several verbs with similar argument structures can productively appear in such related verb frames, this is called an \textit{argument structure alternation}. Examples are listed in Table \ref{tab:overview}. 

For some alternations, there are examples of verbs that participate in both frames (e.g., are positive examples for both dative and double object frames), only the first frame (e.g., are positive examples for the dative frame, but negative examples for the double object one), or only the second frame (e.g., are positive examples for the double object frame and negative ones for the dative frame). However, full empirical coverage is not always possible for every alternation. In our corpora, two of our alternations (\textsc{Causative--Inchoative} and \textsc{\textit{there}-Insertion}) are sparse; some of their frames cannot be provided with negative examples. We discuss this issue in more detail in Section \ref{subsec:LexCorpus}).

\section{Datasets} 
In this section, we describe in detail our word-level dataset, which we call the \textbf{L}exic\textbf{a}l \textbf{V}erb--frame \textbf{A}lternations dataset (LaVA); and the corresponding sentence-level dataset, which we call the \textbf{F}rames and \textbf{A}lternations of \textbf{V}erbs \textbf{A}cceptability dataset (FAVA). Five argument structure alternations are chosen and verbs that evoke at least one frame of the alternation are included in our lexical corpus. These verbs are subsequently used to semi-automatically create a sentence acceptability corpus for our second experiment. We describe our selected argument structure alternations in the remainder of this section and introduce our corpora.

\subsection{LaVA---The Lexical Corpus}\label{subsec:LexCorpus}
We construct LaVA from 515 verbs manually mined from five of the largest syntactic verb frame alternations provided by \newcite{levin1993}: \textsc{Causative--Inchoative, Dative, Spray--Load, \textit{there}-Insertion}, and \textsc{Understood-Object}. Each alternation consists of two different syntactic frames. Our dataset lists whether each verb participates in each frame (wherever available, see the subsection on sparsity below); the alternations and their verb frames are described in the following.

\paragraph{\textsc{Causative--Inchoative}~Alternation~}
The \textsc{Causative--Inchoative} \cite{sunden1916,fillmore1966, hale1986,hale2002} dataset is an expanded version of the \textsc{Causative--Inchoative} dataset from \citet{warstadt2018}, and it contrasts verbs which can evoke both causative and inchoative frames, like \textit{drop} in Table \ref{tab:FrameExamples}, with verbs that can evoke only the causative frame, like \textit{blow}. Importantly, the causative frame is \textit{transitive}---taking two syntactic arguments---and the inchoative frame is \textit{intransitive}---taking only one. In the causative frame, the subject (e.g., \textit{Jessica}) causes the object (e.g., \textit{the vase}) to undergo a change of state (e.g., to be \textit{dropped}), but, in the inchoative frame, the argument which undergoes a change of state is the subject. 

\paragraph{\textsc{Dative}~Alternation} The \textsc{Dative} \cite{bresnan1980, marantz1984,larson1988} dataset consists of verbs that indicate transfer of possession; both frames evoked by these verbs take three arguments, 
but the two frames differ in the order of arguments. In the prepositional dative frame, the \textit{theme} is the syntactic object of the verb, and the \textit{recipient} is within a prepositional phrase; in the dative double object frame, there is no prepositional phrase, and both the \textit{theme} and the \textit{recipient} appear after the verb. Table \ref{tab:FrameExamples} provides examples from the three sets of verbs: one set of verbs evokes both the prepositional dative frame and the double object frame (e.g., \textit{give}), another set only evokes the prepositional dative frame and not the double object frame (e.g.,  \textit{administered}), and the last set of verbs only evokes the double object frame, but not the prepositional dative frame (e.g., \textit{charged}). 

\paragraph{\textsc{Spray--Load}~Alternation} The \textsc{Spray--Load} \cite{tenny1987,levin1995,arad2006} dataset includes transitive verb frames that relate to putting objects in places or covering things with other objects as described in Section \ref{sec:FrameAlternations}.

\paragraph{\textsc{\textit{There}-Insertion}~Alternation}
The \textsc{\textit{there}-Insertion} \cite{poutsma1904, milsark1974, szabolcsi1986} dataset contains intransitive verbs that can evoke a frame in which the subject of the sentence (e.g., \textit{fear}) follows the verb (as in \textit{There remained fear in my mind.}, despite the fact that it would usually appear before the verb in other frames; for these sentences the subject position is filled with a dummy word, \textit{there}. The \emph{there} frame requires a prepositional phrase adjunct---e.g., \textit{There remained fear *(in my mind)}---but the no-\textit{there} frame does not---e.g., \textit{Fear remained (in my mind)}. Verbs that evoke both frames are verbs of existence, spatial configuration, meandering movement, manner of motion, appearance, and inherently directed motion.

\paragraph{\textsc{Understood-Object}~Alternation} The \textsc{Understood-Object} \cite{rice1988,levin1993} dataset contains verb frames that vary in transitivity and describe conventionalized movements of body parts. In the transitive \textsc{Understood-Object} reflexive frame, the body part is the object of the verb (e.g., \textit{Ada clapped her hands.}). In the intransitive \textsc{Understood-Object} no-reflexive frame, the affected \textit{theme} participant (e.g., the body part, or \textit{hands}) is recoverable from the verb (e.g., \textit{clapped}) even though the frame does not require the \textit{theme} (i.e., we know that Ada is clapping her hands and not something else when we interpret the object-less sentence \textit{Ada clapped}).

\paragraph{Sparsity}\label{sparsity} Due to the nature of verb argument structure alternations, in some cases no negative examples can be obtained. For instance, there are no English verbs that can appear in the inchoative, but not the causative (see the first two columns of Table \ref{tab:overview}). This means that, for the \textsc{Causative--Inchoative} alternation, verbs can either evoke both causative and inchoative frames (i.e., be positive examples for both frames) or just the causative frame (i.e., be a positive example for causative and a negative example for inchoative). Similarly, there are verbs that can appear in only no-\textit{there}, but no verbs that can only appear in the \textit{there} frame. This leads to sparsity of annotations. As a result, word-level classifications for these frames are trivial.

Another factor that contributes to data sparsity is that our lexical corpus relies on verbs that \newcite{levin1993} provides as positive (i.e., \textit{grammatical}) or negative (i.e., \textit{ungrammatical}) examples; it does not provide grammaticality judgments for each verb in every frame. In some cases, this is for a linguistic reason: \textsc{Causative--Inchoative} alternation verbs can take at most two arguments, and thus do not appear in frames requiring 3 arguments like the prepositional dative or double object frames. In other cases, there is no obvious reason for a particular verb to not appear in another frame, but the annotations in \citet{levin1993} do not provide that verb--frame combination. In many of these cases, we augment Levin's judgments with our own, also semi-automatically, in attempts to alleviate this issue. However, despite these efforts, the resulting dataset is still sparse, i.e., it does not list whether every verb is a positive or negative example for every frame.

\begin{table}[t]
\small
\centering
\setlength{\tabcolsep}{8.3pt}
\begin{tabular}{l r r}
\toprule
{\bf Levin Class}  &\bf Sentences & \bf {\% Positive} \\
\midrule
\textsc{Causative--Inchoative}  &1168 & 78.9 \\
\textsc{Dative}  &644 & 70.2 \\
\textsc{Spray--Load}  &5127 & 58.6 \\
\textsc{\textit{there}-Insertion} & 718 & 77.0 \\
\textsc{Understood-Object} &  705 & 54.2 \\
\bottomrule
\end{tabular}
\caption{\label{tab:corpuscount}Sentence counts for our acceptability corpus. 
``\% Positive'' is the percentage of sentences that count as acceptable, i.e., as positive examples.}
\end{table}

\subsection{FAVA---Acceptability Judgments Corpus}\label{subsec:AccCorpus}
FAVA is a set of nearly 10k sentences with acceptability judgments. It is constructed semi-automatically from the verbs in the lexical corpus; Table \ref{tab:corpuscount} provides a brief overview. 

Two of the authors, both trained as linguists, manually construct lexical sets consisting of verbs with similar frame-selectional properties that are paired with semantically plausible nouns (and prepositions, where needed). These lexical sets are used to automatically generate sentences with different syntactic frames. For example, the lexical set in \ref{sets} is used to generate 18 minimal pairs of sentences as in \ref{minimal} (one pair for each combination of verb, patient, location, and preposition).

\ex.\label{sets} verbs = \{hung, draped\}\\
patients = \{the blanket, the towel, the cloth\}\\
locations = \{the bed, the armchair, the couch\}\\
prepositions = \{over\}

\ex.\label{minimal}
\a.Betty draped the blanket over the couch.
\b.* Betty draped the couch with the blanket.

\noindent A similar, semi-automatic sentence creation method focusing only on the passive alternation (and non-argument structure syntactic reorderings using negation and relative clauses) was employed by \newcite{ettinger2016} and \newcite{warstadt2018}. 

Using this method, we construct five sentence-level datasets highlighting different verb alternations (\textsc{Causative--Inchoative},\footnote{The \textsc{Causative--Inchoative} dataset presented here is an expanded version of an analysis dataset in \newcite{warstadt2018}.} \textsc{Dative}, \textsc{Spray--Load}, \textsc{\textit{there}-Insertion}, \textsc{Understood-Object}) that are chosen so that sentences could be generated with the maximum of variability in the choice of verbs. 
We split our data into training, development, and test sets by binning lexical sets into training and evaluation bins randomly, in equal proportions. The evaluation set is then split 80/20 into test and development set. Splitting by lexical bin rather than by sentence prevents models from finding a trivial solution to classification by learning to recognize specific verbs and verbal arguments from the training set in the evaluation or test set.

\section{Pre-Trained Representations}\label{sec:pre-trained}
Embeddings, i.e., vector representations of linguistic objects like characters, words, or sentences,
encode helpful information for downstream applications \cite{mikolov-yih-zweig:2013:NAACL-HLT}. In particular, they can be used to leverage knowledge from one task for another and have been shown to improve performance on a diverse set of tasks. Embeddings are usually low-dimensional; common sizes differ between 100 and 300.
Our experiments make use of three types of word and sentence embeddings, which we will describe in the following.

\paragraph{Word Embeddings} 
For our word-level experiments, we use two different embeddings which differ in the way of their creation. First, we use 300-dimensional GloVe embeddings trained on 6B tokens \cite{pennington-14}.\footnote{\url{http://nlp.stanford.edu/data/glove.6B.zip}} 
GloVe embeddings are used frequently in natural language processing (NLP), so evaluating them for knowledge of verb frames will be relevant for their application to and future research on tasks requiring rich syntactic features. Second, we use embeddings trained on the smaller 100M token British National Corpus\footnote{\url{http://www.natcorp.ox.ac.uk}} (BNC), optimizing a language modeling objective. The language model (LM) is a (single-directional) LSTM trained by \newcite{warstadt2018} using PyTorch and optimized using Adam \cite{kingma2014adam}. The BNC data is tokenized using NLTK \cite{bird2004nltk} and words outside the 100k most frequent words in the BNC are replaced with \texttt{<unk>}. 

Our peripheral interest in how humans learn lexical frame-selectional properties motivates us to investigate these LM-trained word embeddings. We reduce the potential differences between human learners and our models by considering embeddings that are trained on an amount of data similar to what humans are exposed to during language acquisition.
For this reason, most publicly available, pre-trained word vectors are a rather unnatural fit, since these embeddings are usually trained on several orders of magnitude more data than humans see in a lifetime.\footnote{If we extrapolate from data gathered by \newcite{hart-92}, we can estimate that children are exposed to about 100 million tokens, on average, by age 5.}

\paragraph{Sentence Embeddings} We further produce sentence embeddings with the help of an existing sentence encoder. Namely, we employ the sentence encoder trained by \newcite{warstadt2018} which performs best in their downstream acceptability classification task. The encoder is trained on a real/fake discrimination task. This is a binary classification task in which a model learns to distinguish \textit{naturally occurring} sentences in the BNC from \textit{fake} sentences. Fake sentences themselves are either generated by a LM 
or by permuting naturally occurring sentences. The real/fake dataset consists of about 12M sentences, including about 6M sentences from the BNC, about 3M million LM-generated sentences, and 3M permuted sentences. The data is tokenized and unknown words replaced in the same way as in the LM training data. A development set 
is used for early stopping. 20 real/fake encoders are trained for 7 days or until the completion of 4 training epochs without improvement in Matthews correlation coefficient on the development set.

\begin{figure}[t]\centering
\includegraphics[width=0.83\columnwidth]{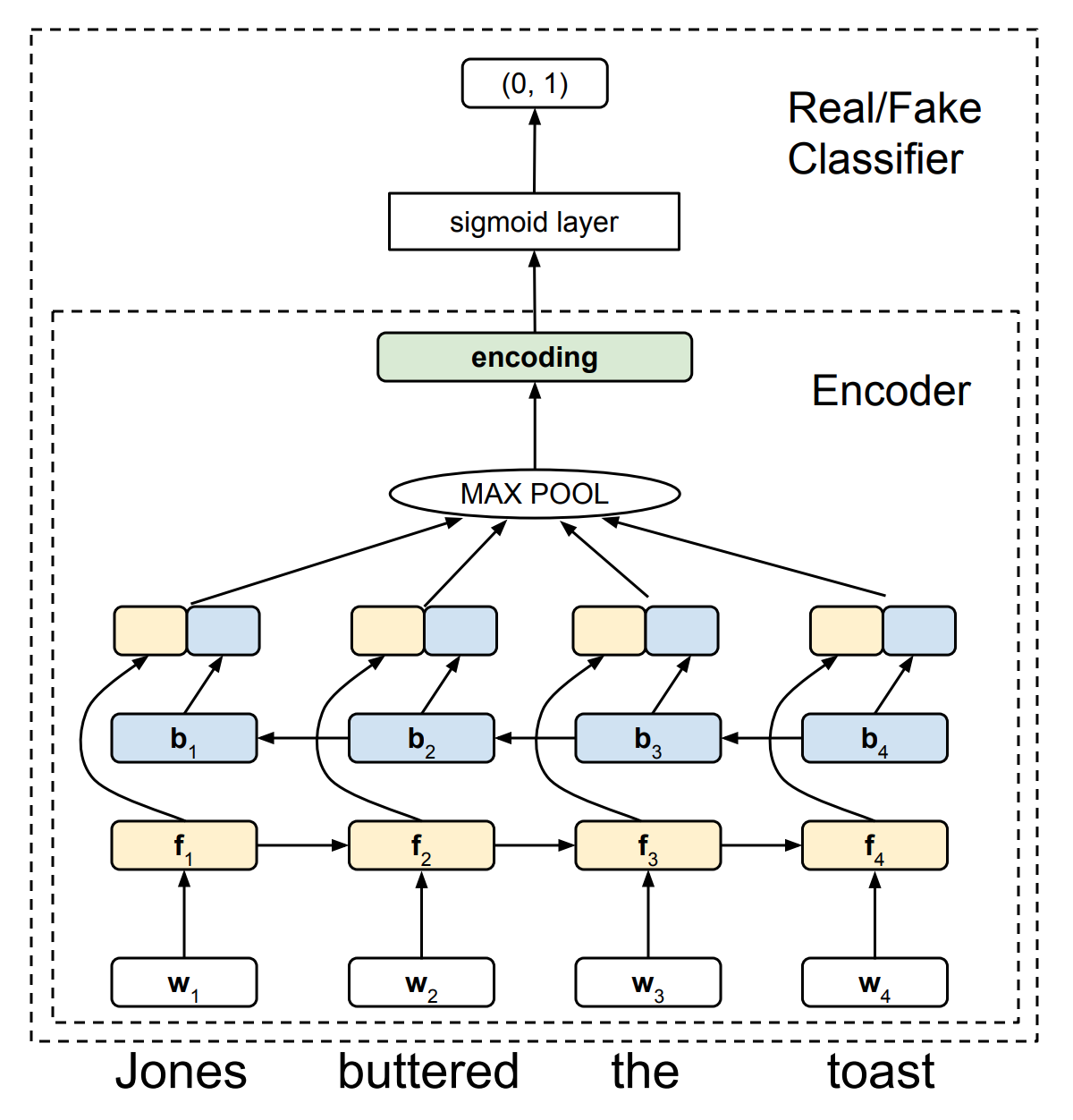}
\caption{Real/fake model. 
$w_i$ = word embeddings, $f_i$ = forward LSTM hidden state, $b_i$ = backward LSTM hidden state. Figure from \newcite{warstadt2018}.}\label{fig:encoder}
\end{figure}

The architecture of the real/fake encoder is shown in Figure \ref{fig:encoder}. A bidirectional long-short term memory network \cite[LSTM,][]{hochreiter1997long} reads the words of a sentence. A fixed-length sentence embedding is then produced by a max-pooling operation over the concatenations of the forward and backward hidden states at each time-step. 
This encoding serves as input to a sigmoid output layer, which outputs a binary prediction.
The input to the encoder 
are ELMo-style \cite{peters-18} contextualized word embeddings from a trained LM. As in ELMo, the representation for a word $w_i$ is a linear combination of the hidden states $h_i^j$ for each layer $j$ in an LSTM LM, though we depart from that paper by using only a forward LM.

As argued in \citet{warstadt2018}, this sentence encoder is a reasonable model for a human learner because it is not exposed to any knowledge of language that could not plausibly be part of the input to a human learner. Its training data consists of the same 100 million tokens used to train the word embeddings, augmented with another 100 million generated tokens in the \emph{fake} data.

\section{Experiment 1: From Word Embeddings to Argument Structures}
In our first experiment, we aim at classifying acceptable syntactic frames, given embeddings for each of the verbs.

\subsection{Model}
\paragraph{Architecture }
We cast the identification of syntactic frames in which a verb can appear as a multi-label classification problem. We train one classifier per alternation, and the classes to be predicted correspond to the participating frames (cf. Table \ref{tab:FrameExamples}), i.e., each classifier predicts values for 2 different classes. 

Employing a multi-layer perceptron (MLP) with a single hidden layer, the probability of a syntactic frame $s$ being acceptable for a given verb is modeled as: 
\begin{align}
p(s) = \sigma(W_2 (f(W_1x))
\end{align}
Here, $x$ is the input, i.e., a word embedding representing a given verb, $W_1$ and $W_2$ are weight matrices, $\sigma$ denotes the sigmoid function, and the activation function $f$ is a rectified linear unit (ReLU). 

\paragraph{Hyperparameters and Training Regime}
We employ the same hyperparameters for all word-level classifiers. 
In particular, we use $30$-dimensional hidden states; note that the size of the embedding vectors is defined by the type of embeddings we use.
During the final classification, we use a threshold of $0.7$ to map the model's predictions to binary outputs.

For training, we use the Adam \cite{kingma2014adam} optimizer. All ANNs are trained for $15$ epochs, but we apply the best performing model on the test set. Further, we use 4-fold cross-validation: the set of verbs is split into 4 equally sized parts out of which 2 are chosen to be the training set, 1 functions as the development set and 1 as the test set. 

\subsection{Metrics}
We report both accuracy and Matthews correlation coefficient \cite[MCC,][]{matthews-75} for this and the following experiment (cf. Section \ref{sec:exp2}), but primarily rely on MCC for evaluation following \newcite{warstadt2018}. MCC is a special case of Pearson's \emph{r} for binary classification. It measures correlation between two binary distributions in the range from -1 to 1, with any two unrelated distributions having a score of 0, regardless of class imbalance. As such, this metric is more robust to unbalanced classification than traditional metrics like F1 or accuracy, both of which favor classifiers with a majority class bias. 

\subsection{Results}
\begin{table*}[ht] 
  \setlength{\tabcolsep}{1.pt}
  \small
  \centering
  \begin{tabular}{ ll cc  cc  cc  cc  cc }
    \toprule
    & & \multicolumn{2}{c }{\textsc{Causative--Inchoative}} & \multicolumn{2}{c }{\textsc{Dative}} & \multicolumn{2}{c }{\textsc{Spray--Load}} & \multicolumn{2}{c }{\textsc{\textit{there}-Insertion}} & \multicolumn{2}{c }{\textsc{Understood-Object}} \\
    &  &  Inch. & (Caus.) & Prep. & 2-Obj. & \textit{with} & Loc. & no-\textit{there} & (\textit{there}) & Refl. & Non-Refl.\\
    \midrule
    \multirow{1}{*}{\textit{CoLA:} Majority BL} 
    & Acc. & 66.7 & (100.0) & 85.0 & 84.9 & 71.0 & 73.9 & 78.7 & (100.0) & 97.7 & 83.0  \\
    \midrule
    \multirow{2}{*}{\textit{CoLA:} MLP} & MCC & \bf 0.555 & 0.0 & 0.32 & \bf 0.482 & \bf 0.645 & 0.253 & \bf 0.459 & 0.0 & 0.0 & 0.219  \\
    & Acc. & 81.0 & (100.0) & 86.6 & 88.3 & 85.8 & 72.9 & 84.3 & (100.0) & 97.7 & 79.0  \\
    \midrule
    \multirow{1}{*}{\textit{GloVe:} Majority BL} 
    & Acc. & 66.8 & (100.0) & 85.0 & 85.3 & 71.0 & 74.6 & 79.1 & (100.0) & 97.6 & 81.5 \\
    \midrule
    \multirow{2}{*}{\textit{GloVe:} MLP} & MCC & \bf 0.672 & 0.0 & 0.0 & 0.0 & \bf 0.585 & 0.145 & \bf 0.536 & 0.0 & 0.0 & 0.3 \\
    & Acc. & 85.5 & (100.0) & 85.0 & 85.3 & 83.9 & 73.4 & 85.8 & (100.0) & 97.6 & 73.2 \\
    \bottomrule
  \end{tabular}
  \caption{\label{tab:results1}Results from Experiment 1 for CoLA-style embeddings (top)
  and GloVe embeddings (bottom); ``Majority BL'' denotes the majority baseline. Bolded MCC values represent reasonably strong correlations (above 0.45). Results for the majority baselines differ due to different words not having a vector representation within the respective embeddings.
  The corpus does not contain negative examples for caus.~and \textit{there} frames (parenthetical); these results cannot be interpreted and are only included for completeness.}
\end{table*}
Table \ref{tab:results1} shows our results.
Our first observation is that, overall, accuracies for GloVe and CoLA-style embeddings are comparable for all classes. This suggests that they both contain similar information about verbs and syntactic frames, and is in line with the fact that both embeddings are based on co-occurrences of words. 

Second, we find that, for GloVe embeddings, the MLP performs on par with the majority baseline for some verb frames, namely causative and \textit{there}, as well as \textsc{Dative} prep. and \textsc{Dative} 2-Obj.; a look at the model predictions reveals that it indeed predicts the majority class for all examples. In this case, MCC will be zero, which is indicative of situations where the model predictions are no better than random. We will not further analyze these cases, since the results  likely indicate that our lexical dataset does not contain enough examples for the model to learn from, and, thus, do not tell us anything meaningful about the embeddings. We would like to note that methods which explicitly account for skewed datasets might help for \textsc{Dative} prep. and \textsc{Dative} 2-Obj., 
but we leave an investigation of such methods for future work.

Finally, we obtain a weak (0.1--0.5) to moderate (0.5--0.7) MCC for both embedding methods and all other classes (with the MLP's accuracy also often being higher than that of the majority baseline). This indicates that information about the evoked syntactic frames can indeed be extracted from verb embeddings. Relatively good performance (${>}0.45$) is found for the inchoative frame (both embeddings), the \textsc{Dative} 2-Obj. frame (CoLA), the \textit{with} frame (both embeddings), and \textit{no}-there frame (both embeddings). Since our classification method (an MLP) is rather simple, our results can be considered a lower-bound on performance, thus showing that verb-frame information is rather obvious in our investigated embeddings.

\section{Experiment 2: From Acceptability to Acceptable Argument Structures}\label{sec:exp2}

Linguists are able to arrive at a classification of a verb according to its syntactic frames by interrogating whether sentences with a given verb and frame are acceptable. Analogously, we can observe whether a verb's frame-selectional properties can be extracted from a sentence embedding by training an acceptability classifier to distinguish sentences with acceptable from sentences with unacceptable verb-frame combinations. If a classifier is able to reliably classify all minimal pairs
of several verbs with different frame-selectional properties from a sentence embedding alone, we can infer that the sentence embedding contains enough information to distinguish both the frame-selectional properties of the verbs and the relevant syntactic frames.

\paragraph{Model}
Our acceptability classifier is again an MLP with a single hidden layer. We model the probability that a that sentence $S$ is acceptable as:

\begin{align}
p(S) = \sigma(W_2 (tanh(W_1x))
\end{align}

Here $x$ is the input, a sentence embedding obtained from the real/fake sentence encoder described in Section \ref{sec:pre-trained}, $W_1$ and $W_2$ are weight matrices, $\sigma$ denotes the sigmoid function, and $tanh$ is the hyperbolic tangent activation function. We use a threshold of 0.5 to map the model's predictions to binary outputs.

\paragraph{Training Details} To select hyperparameters, we train 20 acceptability classifiers on each of the five datasets, and an additional 20 classifiers on a dataset produced by aggregating all the datasets. We repeat all experiments augmenting each dataset with the more than 10k sentences from the corpus of linguistic acceptability (CoLA) built by \newcite{warstadt2018}. Hyperparameters are chosen by random search within the following ranges: hidden size $\in [20, 100]$, learning rate $\in [10^{−2}, 10^{−5}]$, and dropout rate $\in \{0.2, 0.5\}$. All models are trained using early stopping with a patience of 20 epochs.

\begin{table*}[ht]
\setlength{\tabcolsep}{1.5pt}
\centering
\small
\begin{tabular}{l l c c c c c c}
\toprule
							&			& Comb. 	& \textsc{Causative--Inchoative} 	& \textsc{Dative} 	& \textsc{Spray--Load} 	& \textsc{\emph{there}-Insertion} 	& \textsc{Understood-Object} \\\midrule
\multirow{2}{*}{w/o CoLA} 	& MCC 		& 0.290 	& \textbf{0.603}		& 0.413 	& 0.323 		& \textbf{0.528}	& \textbf{0.753}\\
			 				& Acc.		& 64.6 		& 85.4			& 76.0		& 66.2			& 72.9		& 87.4 \\\midrule
\multirow{2}{*}{w/ CoLA} 	& MCC 			& 0.361 	& \textbf{0.464}		& 0.329 	& 0.261 		& \textbf{0.523}	& \textbf{0.638}\\
							& Acc. 		& 68.7		& 81.2			& 59.0		& 63.4			& 72.5		& 81.8 \\\midrule
\multirow{2}{*}{Majority BL} &MCC	&0.0	&0.0	&0.0	&  0.0	&0.0	&0.0
\\ &Acc.	&66.6	&77.6	&82.1	&  60.3	&77.5	&53.7	

\\
                     
                            \bottomrule
\end{tabular}
\caption{\label{tab:acceptabilityresults}Results from Experiment 2. ``w/o CoLA'' are models trained on datasets not augmented with CoLA; ``w/ CoLA'' are models trained on augmented datasets; ``Comb.'' refers to an aggregate dataset. Bolded MCC values represent moderate correlations (above 0.45).}
\end{table*}
\subsection{Results}
Table \ref{tab:acceptabilityresults} shows results for acceptability classification on the verb--frame datasets. These results lead us to conclude that the sentence encoder we test does reliably encode some fine-grained lexical information, but fails to do so in all cases. Our models are able to perform reliable acceptability classifications on several of the alternations featured in FAVA, achieving a moderate correlation (0.5--0.7) in 5 out of 12 experiments, and a strong correlation ($>$0.7) in one experiment. Most classifiers achieve a 
correlation above $0.3$.

Across all 
verb classes, augmenting the training data with CoLA examples lowers MCC. However, when evaluating on the aggregate dataset augmenting the training data with CoLA improves MCC.
One explanation for this might be that the distribution from which the test set is drawn does not resemble the training distribution: for instance, in the \textsc{Causative--Inchoative} with CoLA set, training examples illustrating the relevant alternation are outnumbered about 20:1 by CoLA examples that illustrate mostly unrelated syntactically or semantically complicated phenomena. 

On the other hand, augmenting the combined dataset with sentences from CoLA helps. Performing well on the combined dataset requires an acceptability classifier with knowledge of several unrelated phenomena, so it is not surprising that augmenting the verb-alternation sentences with domain-general CoLA data improves performance. 

The easiest phenomenon by a wide margin for acceptability classifiers was the \textsc{Understood-Object} alternation. One explanation for this fact might be that the semantic relatedness of verbs like \emph{blink} and objects like \emph{her eyes} makes it easier to recognize from the sentence embedding whether their co-occurrence is expected or anomalous; for example, \textit{eye} is the most common collocate for \textit{blink}, \textit{hand} is the most common one for \textit{clap}, and \textit{tooth} is in the top five most common collocates for \textit{chip} \citep{davies2008,davies2009}.\footnote{\url{https://corpus.byu.edu/coca/}}

The next easiest alternations for our models to learn are \textsc{Causative--Inchoative} and \textsc{\emph{there}-Insertion}, both of which have at least one intransitive verb frame (both frames are intransitive in the case of \textsc{\textit{there}-Insertion}, but in one frame there is a locative adjunct). One common denominator among these three easiest alternations for the acceptability model is that they all involve verbs appearing in an intransitive frame (in the case of \textsc{\emph{there}-Insertion} a locative adjunct is present as well). By contrast, the \textsc{Dative} and 
\textsc{Spray--Load} alternations both involve verbs that take multiple arguments, appearing with up to three arguments (or possibly two arguments and a locative adjunct) in all frames. Intransitive verb frames are the simplest syntactic frames possible, and it might be expected that they are easiest to recognize. 

Qualitatively, we do not find that the amount of training examples in the dataset was correlated with performance. By way of illustration, the \textsc{Spray--Load} alternation accounts for over half of all the generated data, yet it was by far the hardest individual alternation for our models to learn.

\section{Related Work}
This investigation is part of a growing body of work which seeks to investigate the linguistic competence of ANNs. For instance, a study by \newcite{linzen-16} tested the ability of ANNs to identify mismatches in subject--verb agreement, even in the presence of intervening ``distractor'' nouns. Similarly, \newcite{ettinger2016} investigated whether sentence embeddings contain grammatical information, e.g., about the syntactic scope of negation.

Further previous studies on which types of information are contained in embeddings include \newcite{bjerva-augenstein:2018:N18-1}, which asked whether certain phonological, morphological and syntactic information can be extracted from language embeddings. \newcite{malaviya-neubig-littell:2017:EMNLP2017} predicted features from language embeddings which were trained as part of an ANN for machine translation.
Finally, \newcite{ostling-tiedemann:2017:EACLshort} learned language embeddings via multilingual language modeling and used them to reconstruct genealogical trees.
However, 
we are interested in \textit{word} or \textit{sentence} embeddings.
Extracting information from word embeddings is a common task in natural language processing. While most NLP research is application-oriented and directly or indirectly focuses on obtaining embeddings which contain as much knowledge about the task at hand as possible (e.g., by varying the training corpus or embedding method), we are interested in the question how much information is trivially contained in selected popular embeddings.

Also worth mentioning here is a lexical resource named VerbNet \cite{kipper2005,kipper2006}. 
This database contains verbs which were classified according to their semantic and syntactic properties, including their Levin classes.\footnote{To be exact, the set of classes was extended to a superset of the original Levin classes.}
VerbNet has been used in various NLP applications, e.g., semantic role labeling \cite{giuglea-moschitti:2006:COLACL}, word sense disambiguation \cite{Brown:2011:VCA:2002669.2002679}, information extraction \cite{mausam-EtAl:2012:EMNLP-CoNLL}, or investigation of human language acquisition \cite{Korhonen3621}. While this resource is very extensive, it only provides a few example sentences (generally only one or two per frame) for each verb. Since we want to investigate if argument structure information is present in sentence embeddings, we create a larger 
corpus.

\section{Conclusions}
We present complementary word-level and sentence-level datasets, LaVA and FAVA, covering five verb-alternations. 
We train classifiers on verb embeddings to distinguish which syntactic frames a verb can evoke and which it cannot. We further train acceptability classifiers with sentence embeddings as input for sentences which do or do not contain acceptable verb--frame combinations. 
We conclude that information about verb-argument structure alternations is present in both word-level and sentence-level embeddings.
However, some frames seem to be easier to judge than others, and for only few frames a strong correlation can be obtained between model predictions and our gold annotations. There is considerable opportunity for future work which generalizes these experiments to other sentence encoders, verb alternations, and lexical properties.

\section*{Acknowledgments}
This project has benefited from financial support to SB and KK from Samsung Research, and to SB from Google.

\bibliography{naaclhlt2018}
\bibliographystyle{acl_natbib}

\end{document}